\documentclass[11pt]{article}
\usepackage[]{acl}
\usepackage{times}
\usepackage{latexsym}
\usepackage{booktabs}
\usepackage{graphicx}
\usepackage{xcolor}
\usepackage{amssymb}
\usepackage[T1]{fontenc}
\usepackage[utf8]{inputenc}
\usepackage{microtype}

\newcommand{\FOneNumRaces}{157}
\newcommand{\FOneNumInstances}{7253}
\newcommand{\FOneNumSeasons}{8}
\newcommand{\FOneFirstSeason}{2018}
\newcommand{\FOneLastSeason}{2026}
\newcommand{\FOneNumTrain}{6004}
\newcommand{\FOneNumTest}{1249}
\newcommand{\FOneNumTestSample}{207}
\newcommand{\FOneNumStint}{1500}
\newcommand{\FOneNumUndercut}{2034}
\newcommand{\FOneNumOvercut}{1125}
\newcommand{\FOneNumDefense}{2444}
\newcommand{\FOneNumRaceSummary}{150}
\newcommand{\FOneNumStints}{8486}
\newcommand{\FOneNumPits}{5358}
\newcommand{\FOneNumBattles}{3337}

\IfFileExists{result_macros.tex}{\newcommand{\JudgePearson}{0.55}
\newcommand{\JudgeSpearman}{0.54}
\newcommand{\JudgeN}{120}
\newcommand{\JudgeModel}{gpt-5.5}
\newcommand{\MethodFirst}{0.640}
\newcommand{\MethodFinal}{0.881}
\newcommand{\MethodModel}{gpt-5.4-mini}

\newcommand{\ExtractorEnSpearman}{0.80}
\newcommand{\ExtractorEnPearson}{0.50}
\newcommand{\ExtractorEnN}{564}

\newcommand{\XfamSpearman}{1.00}
\newcommand{\XfamPearson}{0.82}
\newcommand{\XfamN}{1090}
\newcommand{\XfamModel}{DeepSeek-V3.2}
\newcommand{\CovLang}{PT}
\newcommand{\CovFlipModel}{grok-4.3}
\newcommand{\CovFlipPrec}{0.89}
\newcommand{\CovFlipRecall}{0.46}
\newcommand{\CovFlipFone}{0.61}
\newcommand{\CovFlipFRank}{last}
\newcommand{\CovTopFoneModel}{claude-sonnet-4-6}
\newcommand{\CovTopFone}{0.83}

\newcommand{\AblMeanRecallA}{0.60}
\newcommand{\AblMeanRecallB}{0.47}
\newcommand{\AblNModels}{5}
\newcommand{\AblNUp}{2}
\newcommand{\RSLoPrecModel}{DeepSeek-V3.2}
\newcommand{\RSLoPrec}{0.49}
\newcommand{\RSLoRecall}{0.42}

}{}

\title{Precision Is Not Faithfulness: Coverage-Aware Evaluation of\\
Grounded Generation with a Complete Oracle}
\author{Juan S. Santillana\thanks{The author is a DevOps Engineer at Globant;
  academic affiliation pending.} \\
  \texttt{juan.salas@globant.com}}

\begin{document}
\maketitle

\begin{abstract}
Reference-free faithfulness metrics verify each atomic claim a model makes against
ground truth, and are increasingly used to evaluate grounded generation. We show they
share a blind spot: they measure only \emph{precision} -- are the stated claims
supported? -- and therefore \emph{reward abstention}, since a model can score
near-perfect faithfulness by saying almost nothing. We make this measurable using
Formula~1 telemetry, a domain where strategic ground truth is derived deterministically
and, crucially, \emph{completely}: for each decision we know the full set of facts that
mattered. This completeness -- absent in open-domain faithfulness benchmarks -- lets us
measure \emph{recall} (coverage of the relevant facts) alongside precision. On a
multilingual (EN/ES/PT) benchmark of \FOneNumInstances{} decision instances spanning
\FOneNumRaces{} races, the most precise frontier model (\CovFlipModel{}, precision
\CovFlipPrec{}) covers only \CovFlipRecall{} of the relevant facts and ranks
\CovFlipFRank{} by $F_1$, so requiring coverage \emph{reorders the systems}; the same
effect reappears in a second complete-oracle domain (weather forecasts). We
pair faithfulness with coverage into a single score, validate the metric (controlled
perturbation; agreement across a model-free regex extractor and a cross-family LLM
extractor, system-level Spearman \XfamSpearman{}), and give a verifier-guided generation
method that improves precision and recall without references. We release the benchmark,
structured annotations, metric, baselines, and an interactive demo.
\end{abstract}

\section{Introduction}
Reference-free faithfulness metrics -- which decompose a generation into atomic claims
and verify each against ground truth
\citep{min2023factscore,fabbri2022qafacteval} -- have become a standard way to evaluate
grounded generation without gold reference texts. They report a \emph{precision}: of the
claims the model made, how many are supported? We argue this is only half of what
``faithful'' should mean. A model can maximize precision by being maximally cautious --
stating one safe fact and omitting everything else. A precision-only metric scores such
output as near-perfect, even though it is uninformative. Faithfulness, measured as
precision alone, \emph{rewards abstention}.

The missing half is \emph{recall}: of the facts that mattered, how many did the model
correctly state? Open-domain faithfulness benchmarks cannot measure this, because there
is no complete, enumerable set of relevant facts to recall against -- the ground truth
is whatever a retriever or annotator happened to surface. We therefore turn to a domain
with a \emph{complete} oracle. Formula~1 produces rich, public timing and telemetry data
from which strategic ground truth (pit laps, compounds, undercuts, defenses, outcomes)
can be derived \emph{deterministically and exhaustively}: for each decision we know the
full set of checkable facts. This lets us measure recall, and pair it with precision,
in a way open-domain settings structurally cannot.

Using this oracle we show the abstention problem is not hypothetical: on a multilingual
benchmark, the frontier model with the highest faithfulness (precision) covers only
\CovFlipRecall{} of the relevant facts, and once coverage is required the model ranking
\emph{changes} (Section~\ref{sec:exp}). We frame F1 strategy explanation as faithful,
data-grounded NLG -- not race-outcome prediction, which is saturated and outside language
research -- because it gives the cleanest available testbed for this measurement question.

\noindent Our contributions:
\begin{enumerate}
  \item We show that reference-free faithfulness metrics reward abstention, and that with
        a \emph{complete} oracle the model ranking inverts when coverage is required --
        replicated across two unrelated domains, F1 and weather
        (Section~\ref{sec:exp}, \ref{sec:weather}).
  \item A complete-oracle, multilingual (EN/ES/PT) benchmark of \FOneNumInstances{}
        grounded F1 decision instances, and a precision+recall faithfulness metric,
        validated by controlled perturbation and by agreement across a model-free and a
        cross-family extractor (Section~\ref{sec:data}, \ref{sec:metric}).
  \item A verifier-guided generation method that improves both precision and recall using
        only the structured verifier as signal (Section~\ref{sec:method}).
\end{enumerate}
\noindent The benchmark, metric, baselines, and an interactive demo are
public.\footnote{Demo: \url{https://huggingface.co/spaces/jsantillana/faithful-strategy-engineer-f1}}

\section{Related Work}
Data-to-text generation has a long line of sports and structured-record work, e.g.\
RotoWire \citep{wiseman2017challenges}, SportSett:Basketball
\citep{thomson2020sportsett}, weather forecasts from records \citep{liang2009weather},
and controlled table-to-text such as ToTTo \citep{parikh2020totto}; in this setting
\citet{thomson2020accuracy} give a manual gold-standard methodology for \emph{accuracy}
that scores each fact in a generated text, the closest prior practice to our automatic
check.

\paragraph{Faithfulness is precision.} Hallucination/faithfulness evaluation -- the
foundational faithful-vs-factual distinction \citep{maynez2020faithfulness}, FActScore
\citep{min2023factscore}, QA-based consistency \citep{fabbri2022qafacteval}, NLI-based
inconsistency \citep{laban2022summac,kryscinski2020factcc}, consistency benchmarks
\citep{honovich2022true}, sampling-based detection \citep{manakul2023selfcheckgpt}, and
attribution/revision \citep{gao2023rarr}; see \citet{ji2023survey} for a survey -- is
\emph{precision}-oriented: it scores the claims a model makes, not the ones it omits, so it
does not penalize an uninformative answer. That is the abstention blind spot we study.

\paragraph{Recall of facts.} The complementary recall side is closest to our work.
Table-to-text PARENT \citep{dhingra2019parent} credits coverage of the source table but
needs a reference text and a (possibly incomplete) table; for long-form open-domain text,
SAFE \citep{wei2024longfact} reports factual precision \emph{and} recall ($F_1@K$), and RAG
evaluation (RAGAS) scores faithfulness against retrieved context \citep{es2024ragas}.
Crucially, these estimate recall against \emph{retrieved or sampled} facts -- an inherently
incomplete denominator -- whereas our oracle is derived deterministically and is
\emph{complete}, giving an \emph{exact} recall denominator rather than an estimate, which
is what makes the precision/recall ranking inversion measurable. Our verifier-guided
generation relates to self-refinement \citep{madaan2023selfrefine}, but the feedback signal
is a deterministic check against structured data, not the model's own critique. Our use of
a compact, domain-specialized model follows recent work on small efficient models with
native tool use for Spanish technical domains \citep{vectrayxnano2026}.

\section{Task}\label{sec:task}
Each instance provides a structured context (driver stints, tyre compound and
age, pit stops, gaps, safety-car/VSC status) and a decision prompt in EN/ES/PT.
The model must explain a strategic decision such that every factual claim is
verifiable against the context. The benchmark spans five decision types: tyre
strategy, undercut, overcut, \emph{on-track defense} (a faster pursuer kept behind
for several laps -- the ``rear gunner'' move), and \emph{race summary} (a grounded
recap of result and key moments).

\section{Dataset}\label{sec:data}
We extract timing and telemetry with FastF1 \citep{fastf1} (the official live-timing
API; no broadcast access required) and derive strategic events with deterministic
rules. Stints are segmented per driver with a within-stint degradation slope fit on
\emph{green-flag} laps only (excluding in/out laps and laps under yellow, safety car
(SC), or virtual safety car (VSC)) so that pace estimates are not contaminated by
neutralizations. Pit stops are recovered from stint transitions and flagged when made
under SC/VSC. Undercuts and overcuts are detected pairwise but kept only for drivers
genuinely racing each other -- within a small on-track gap so the pit sequence, not
accumulated pace difference, dominates the gap swing -- which yields high-precision
events (spot-checked against known race narratives, e.g.\ the winning one-stop at the
2024 Italian GP). On-track defenses are detected as runs of $\geq 5$ consecutive laps
in which a pace-faster pursuer is held within $1.5$s, flagged when a teammate is
protected ahead; this recovers canonical cases (e.g.\ Alonso holding Hamilton for
$11$ laps, Hungary 2021). Season standings (race plus sprint points) support grounded
championship summaries.

Each instance pairs a serialized structured context with a decision prompt in EN/ES/PT
and the structured ground truth used for verification. From \FOneNumRaces{} races
across \FOneNumSeasons{} seasons (\FOneFirstSeason{}--\FOneLastSeason{}) we obtain
\FOneNumStints{} stints, \FOneNumPits{} pit stops, and \FOneNumBattles{} pit battles,
yielding \FOneNumInstances{} instances (\FOneNumStint{} tyre-strategy,
\FOneNumUndercut{} undercut, \FOneNumOvercut{} overcut, \FOneNumDefense{} defense,
\FOneNumRaceSummary{} race-summary). To avoid leakage we split by
season: \FOneNumTrain{} train instances (\FOneFirstSeason{}--2024) and \FOneNumTest{}
held-out test instances (\FOneLastSeason{}); the model comparisons below use a
stratified test sample of \FOneNumTestSample{} instances. We do not redistribute raw
FOM data; we release the derived structured ground truth, annotations, and code
(Section~\ref{sec:ethics}).

\paragraph{A complete oracle.} The key property we exploit is that this ground truth is
not just deterministic but \emph{complete}: for each decision instance we can enumerate
the full set of checkable facts that a good explanation should cover -- for a stint, the
driver's stop count, pit laps, compound changes, and finishing position; for an
undercut/overcut, the two pit laps, the move and its outcome, and the time gained. This
enumerable fact set is the denominator for recall (Section~\ref{sec:metric}), and is
exactly what open-domain faithfulness settings lack.

\section{Metric: Precision and Recall}\label{sec:metric}
We decompose an explanation into typed atomic claims (pit lap, compound change,
stop count, stint compound, final position, undercut/overcut, outcome, time
gained) and verify each against the structured ground truth, labeling it
\emph{supported}, \emph{contradicted} (the context contains the relevant fact and it
differs), or \emph{unverifiable} (the context lacks the fact). The metric is
reference-free: it never compares to a gold text, only to the structured data the model
was given.

\paragraph{Precision (faithfulness).} The supported fraction of the claims a model makes
is its faithfulness, or \emph{precision}; we report the hard-hallucination (contradicted)
rate alongside. This is the standard reference-free faithfulness quantity -- and, on its
own, it rewards abstention: a model that states a single safe fact scores $1.0$.

\paragraph{Recall (coverage).} Because the oracle is complete (Section~\ref{sec:data}),
we also measure \emph{recall}: of the enumerable facts that mattered for an instance, the
fraction the model correctly stated (i.e.\ produced a supported claim for). A terse model
that omits most facts is now penalized. We summarize the two with their harmonic mean
($F_1$). A faithful explanation should be both \emph{accurate} (high precision) and
\emph{informative} (high recall); reporting precision without recall is the blind spot we
quantify in Section~\ref{sec:exp}.

\paragraph{Two extraction backends.} Claim extraction has two interchangeable
backends that share the schema. A dependency-free \emph{regex} extractor (with light
coreference) targets English and is fast and transparent; we use it for the offline
validation. An \emph{LLM} extractor reads free-form output in any language and emits
the same typed claims, which we use to score Spanish and Portuguese fairly and to keep
a single extractor across languages in the cross-lingual comparison. Verification is
identical in both cases. We validate the metric itself two ways. First, by controlled perturbation: a template
that emits only true statements scores a perfect supported fraction (no false
contradictions from the verifier), while injected errors are penalized in proportion
(Table~\ref{tab:pilot}). Second, by correlation with an independent LLM judge
(\JudgeModel{}, distinct from the extractor) over a mixed sample of $\JudgeN{}$
explanations spanning a range of faithfulness: the automatic score correlates
positively with the judge (Pearson $\JudgePearson{}$, Spearman $\JudgeSpearman{}$),
with our metric being the stricter of the two. Human-judge correlation is future
work.

\IfFileExists{example.tex}{\begin{figure}[t]
\centering
\small
\fbox{\parbox{0.93\columnwidth}{
\textbf{gpt-5.4-mini briefing (excerpt).} \textit{VER’s **overcut attempt on RUS did not work as the decisive move**, because the pit timing data shows it was actually **RUS who pitted first**. What the data shows: - **VER pit lap 12** from MEDIUM to HARD, while **RUS pit lap 13** from MEDIUM to HARD. - On that first cycle, VER stayed out **one lap longer** than [...]}\\[4pt]
\textbf{Faithfulness audit.}\\
\textcolor{red}{$\times$ contradicted}: move actually worked\\
\textcolor{teal}{$\checkmark$ supported}: VER pitted lap 12; RUS pitted lap 13\\
Score: 6/8 supported, 1 contradicted.
}}
\caption{A real frontier briefing where the verifier flags an ungrounded claim against the telemetry while confirming the rest.}
\label{fig:example}
\end{figure}
}{}

\section{Method: Verifier-Guided Generation}\label{sec:method}
We generate an explanation, run the verifier, and feed back both contradicted claims
(fix these) and \emph{uncovered} ground-truth facts (add these) as targeted edit
instructions, iterating a few rounds. Because the signal includes the facts the model
omitted -- available only because the oracle is complete -- the loop targets precision and
recall jointly, not just precision. It uses only the structured verifier (no reference
text) and is applicable to any LLM backend.

\section{Experiments}\label{sec:exp}
\paragraph{Metric validation (offline).} We validate the metric with a
controlled-perturbation study: a deterministic template that emits only true
statements vs.\ one with injected factual errors. Table~\ref{tab:pilot} shows the
metric assigns perfect faithfulness to grounded text (no false contradictions)
and sharply penalizes perturbations.

\begin{table}[t]
\centering
\small
\begin{tabular}{lccc}
\toprule
System & Faithfulness & Halluc. & \#Claims \\
\midrule
template\_faithful & 1.000 & 0.000 & 961 \\
template\_noisy & 0.593 & 0.407 & 901 \\
\bottomrule
\end{tabular}
\caption{Pilot faithfulness on the controlled-perturbation validation (207 instances, lang=en). The faithful template scores 1.0 (no false contradictions from the verifier); the perturbed template is correctly penalized.}
\label{tab:pilot}
\end{table}

\begin{figure*}[t]
\centering
\includegraphics[width=0.92\textwidth]{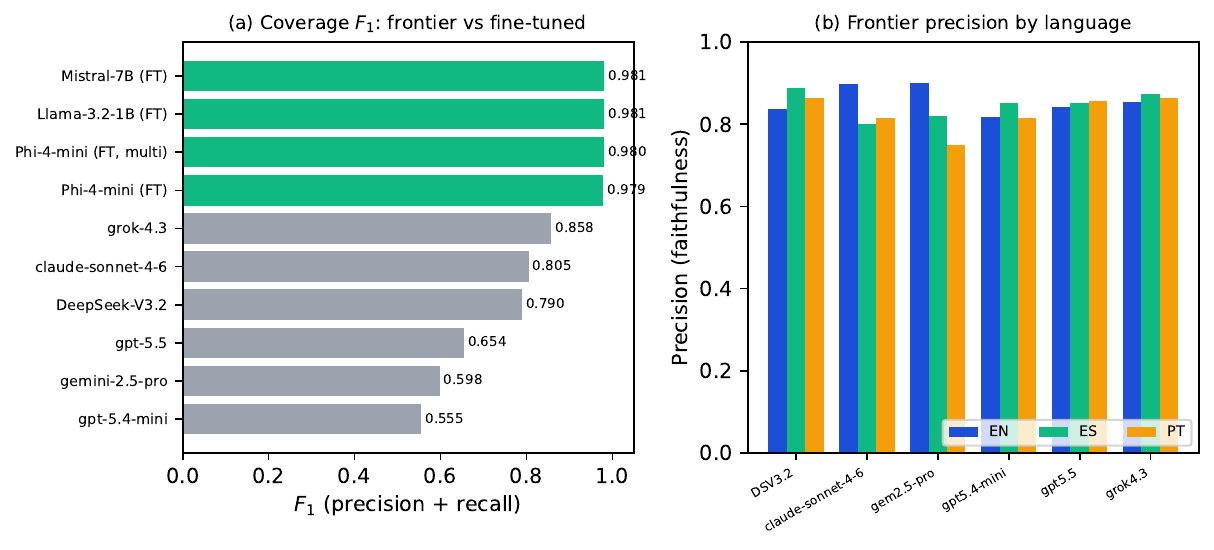}
\caption{(a) Coverage $F_1$ (precision + recall against the complete oracle) on the
held-out 2025 test: fine-tuned small models (green, 1B--7B) reach ${\sim}0.98$, far
exceeding every zero-shot frontier system including Claude Sonnet ($0.805$ EN) and
grok-4.3 ($0.858$ EN). (b) Frontier precision by language (EN/ES/PT); Claude is
the only model whose precision holds or improves outside English.}
\label{fig:results}
\end{figure*}

We study: \textbf{RQ1} -- does precision-only faithfulness reward abstention, and does
requiring coverage change which models look best?; \textbf{RQ2} -- does faithfulness vary
across languages?; \textbf{RQ3} -- can a fine-tuned small model match or exceed the
frontier? We evaluate the latest frontier models from five families zero-shot: OpenAI
gpt-5.5 and gpt-5.4-mini (Azure), xAI grok-4.3, Google gemini-2.5-pro (Vertex AI),
Anthropic claude-sonnet-4-6 (AWS Bedrock), and DeepSeek-V3.2, each scored with the
language-agnostic LLM claim extractor. Even these
models leave a non-trivial fraction of claims ungrounded and produce hard contradictions
(Table~\ref{tab:coverage}): fluent, expert-sounding narratives are not automatically
faithful.

\paragraph{Precision rewards abstention; coverage reorders the ranking (RQ1).}
Table~\ref{tab:coverage} reports precision (faithfulness), recall (coverage), and $F_1$
against the complete oracle. Precise models are not the most complete: in \CovLang{}, the
\emph{most precise} model, \CovFlipModel{} (precision $\CovFlipPrec{}$), covers only
$\CovFlipRecall{}$ of the facts that mattered and so ranks \emph{\CovFlipFRank{}} by $F_1$
($\CovFlipFone{}$), while \CovTopFoneModel{} leads by $F_1$ ($\CovTopFone{}$). Conversely,
more verbose models (e.g.\ claude-sonnet-4-6, DeepSeek-V3.2) are slightly less precise but
far more complete and rise sharply under $F_1$. The ranking by precision and the ranking by $F_1$ disagree
in every language. A precision-only faithfulness metric therefore does not just mis-score
an output, it \emph{reorders} the comparison between systems. This is the paper's central
finding, and it is measurable only because the oracle is complete.

\paragraph{Multilingual (RQ2).} Precision and the precision/$F_1$ disagreement are stable
across EN/ES/PT (Table~\ref{tab:coverage}); the coverage ranking change holds in all three
languages. Most models degrade in non-English: DeepSeek recall drops from $0.76$ (EN) to
$0.50$ (PT); grok from $0.84$ (EN) to $0.46$ (PT). The exception is claude-sonnet-4-6,
whose recall \emph{rises} in ES/PT ($0.83$/$0.84$ vs.\ $0.73$ EN) via more verbose
generations (${\sim}11$ claims/inst vs.\ $8$), making it the most coverage-stable frontier
model across languages. The salient cross-lingual infrastructure effect is the AIServices
safety filter, which blocks the \emph{same} block of English inputs for both AIServices
models (excluded above), with almost no effect in ES/PT -- a reminder that platform-level
filtering, not just the model, shapes what a multilingual evaluation measures.

\paragraph{Small fine-tuned model and verifier-guided method (RQ3).}
Table~\ref{tab:models} compares an open small model (Qwen2.5-3B
\citep{qwen2025}) zero-shot and after LoRA fine-tuning on grounded explanations, on the
test sample under the same precision+recall metric. Read through the lens of RQ1, the
fine-tuned 3B model is the encouraging case: it is both \emph{accurate} and
\emph{complete} -- among the highest $F_1$ in the study -- by faithfully reproducing the
deterministic grounded templates. This is genuine on this distribution but is exactly the
template-mimicry caveat we flag: precision and recall both look near-perfect because the
templates state the key facts, so out-of-template evaluation is the real test (Limitations).
It still echoes evidence that small, specialized models are competitive in focused domains
\citep{vectrayxnano2026}. We replicate this across base families and scales
(Table~\ref{tab:coverage}): LoRA-fine-tuned Llama-3.2-1B, Phi-4-mini and Mistral-7B all
reach $F_1 \approx 0.98$ (precision $\approx 0.99$, recall $\approx 0.97$), each beating
every zero-shot frontier system on $F_1$ while staying substantive ($\sim$5.6
claims/instance, no abstention) -- the gain is not a quirk of one model or of scale (it
holds at 1B). A single \emph{multitask} fine-tune (strategy+commentary) matches its
single-task counterpart ($F_1\,0.980$ vs.\ $0.979$), i.e.\ no measurable task interference.
Separately,
verifier-guided self-correction (Section~\ref{sec:method}) applied to \MethodModel{}
raises precision from $\MethodFirst{}$ to $\MethodFinal{}$ (English regex verifier),
suggesting the structured verifier is a usable training-free signal.

\paragraph{Extractor robustness.} A concern with an LLM-scored metric is that the
extractor (gpt-5.x) shares a family with one evaluated system (gpt-5.5), which could
inflate its score. We re-score the \emph{same} generations with two independent
extractors. (i) A model-free, deterministic \emph{regex} extractor (English): it agrees
with the LLM extractor on the system ranking (Spearman $\ExtractorEnSpearman{}$) and at
the instance level (Pearson $\ExtractorEnPearson{}$, $N=\ExtractorEnN{}$). (ii) A
\emph{cross-family} LLM extractor (\XfamModel{}, no shared lineage with gpt-5.x), across
all three languages: agreement is near-perfect at the system level (Spearman
$\XfamSpearman{}$) and strong per instance (Pearson $\XfamPearson{}$, $N=\XfamN{}$),
higher than the metric's correlation with the independent LLM judge. The same-family
system, gpt-5.5, is not ranked top by its own family's extractor under either check, so
the metric does not favor it.

\paragraph{Open-ended summaries.} The same metric extends to the open-ended
\emph{race-summary} task (recap the result and key moments), which our oracle scores
against winner, finishing positions, battles, and defenses. Here the failure mode flips:
free-form recaps are \emph{verbose but imprecise} -- \RSLoPrecModel{} reaches recall
$\RSLoRecall{}$ at precision only $\RSLoPrec{}$ (many narrative claims the data cannot
confirm) -- so ranking by precision and by $F_1$ again disagree. Precision and recall
capture both failure modes; neither alone does.

\IfFileExists{coverage_table.tex}{\begin{table*}[t]
\centering\small
\begin{tabular}{llcccc}
\toprule
Model & Lang & Prec. & Recall & F1 & Cl./inst. \\
\midrule
DeepSeek-V3.2$^{-71}$ & EN & 0.818 & 0.763 & 0.790 & 9.6 \\
DeepSeek-V3.2$^{-3}$ & ES & 0.841 & 0.719 & 0.776 & 9.3 \\
DeepSeek-V3.2$^{-3}$ & PT & 0.855 & 0.495 & 0.627 & 6.4 \\
gemini-2.5-pro & EN & 0.824 & 0.469 & 0.598 & 5.2 \\
gemini-2.5-pro & ES & 0.845 & 0.451 & 0.588 & 4.9 \\
gemini-2.5-pro$^{-1}$ & PT & 0.855 & 0.483 & 0.617 & 4.6 \\
gpt-5.4-mini & EN & 0.820 & 0.420 & 0.555 & 5.0 \\
gpt-5.4-mini & ES & 0.863 & 0.473 & 0.611 & 4.7 \\
gpt-5.4-mini & PT & 0.850 & 0.493 & 0.624 & 5.3 \\
gpt-5.5 & EN & 0.861 & 0.528 & 0.654 & 5.9 \\
gpt-5.5 & ES & 0.867 & 0.499 & 0.634 & 5.3 \\
gpt-5.5 & PT & 0.886 & 0.511 & 0.648 & 5.4 \\
grok-4.3$^{-71}$ & EN & 0.878 & 0.838 & 0.858 & 8.9 \\
grok-4.3$^{-3}$ & ES & 0.901 & 0.710 & 0.794 & 6.8 \\
grok-4.3$^{-3}$ & PT & 0.887 & 0.462 & 0.608 & 4.3 \\
claude-sonnet-4-6 & EN & 0.897 & 0.730 & 0.805 & 8.2 \\
claude-sonnet-4-6 & ES & 0.801 & 0.832 & 0.817 & 10.6 \\
claude-sonnet-4-6 & PT & 0.815 & 0.844 & 0.829 & 11.7 \\
\midrule
\multicolumn{6}{l}{\emph{Small models LoRA-fine-tuned on the complete oracle (this work)}}\\
\midrule
Llama-3.2-1B (FT) & EN & 0.994 & 0.968 & \textbf{0.981} & 5.6 \\
Phi-4-mini (FT) & EN & 0.994 & 0.965 & \textbf{0.979} & 5.6 \\
Mistral-7B (FT) & EN & 0.994 & 0.967 & \textbf{0.981} & 5.6 \\
Phi-4-mini (FT, multitask) & EN & 0.993 & 0.968 & \textbf{0.980} & 5.6 \\
\bottomrule
\end{tabular}
\caption{Precision (faithfulness) is gameable by abstention: against the \emph{complete} oracle we also measure recall (coverage of the facts that mattered). The most precise model is \emph{not} the most informative; requiring coverage ($F_1$) reorders the systems (e.g.\ in \CovLang{}, the most precise model, \CovFlipModel{}, ranks \CovFlipFRank{} by $F_1$). Only a complete structured oracle makes recall measurable. \textbf{Fine-tuning on the complete oracle closes the gap}: every FT model---even 1B---reaches $F_1 \approx 0.98$, beating every frontier system while remaining substantive. $^{-n}$: $n$ English instances dropped by platform content-filtering (same instances across AIServices models; not model behavior).}
\label{tab:coverage}
\end{table*}
}{}
\IfFileExists{ablation_table.tex}{\begin{table}[t]
\centering\small
\resizebox{\columnwidth}{!}{%
\begin{tabular}{lcccccc}
\toprule
 & \multicolumn{3}{c}{Default prompt} & \multicolumn{3}{c}{Cover-all prompt} \\
\cmidrule(lr){2-4}\cmidrule(lr){5-7}
Model & P & R & F1 & P & R & F1 \\
\midrule
DeepSeek-V3.2 & 0.82 & 0.76 & 0.79 & 0.84 & 0.57 & 0.68 \\
gemini-2.5-pro & 0.82 & 0.47 & 0.60 & 0.85 & 0.56 & 0.68 \\
gpt-5.4-mini & 0.82 & 0.42 & 0.56 & 0.87 & 0.52 & 0.65 \\
gpt-5.5 & 0.86 & 0.53 & 0.65 & 0.89 & 0.43 & 0.58 \\
grok-4.3 & 0.88 & 0.84 & 0.86 & 0.88 & 0.27 & 0.41 \\
\bottomrule
\end{tabular}}
\caption{Prompt-sensitivity ablation (English): the neutral \emph{default} prompt vs.\ an explicit \emph{cover-all} prompt that asks the model to state every supportable fact. Asking for completeness does \emph{not} close the coverage gap (mean recall $\AblMeanRecallA{}$ vs.\ $\AblMeanRecallB{}$; only $\AblNUp{}$ of $\AblNModels{}$ models improve) -- extra verbosity does not add the key facts. The low coverage is therefore not an under-prompting artifact, and precision-only faithfulness reports none of this.}
\label{tab:ablation}
\end{table}
}{}
\IfFileExists{weather_table.tex}{\begin{table*}[t]
\centering\small
\begin{tabular}{llcccc}
\toprule
Model & Lang & Prec. & Recall & F1 & Cl./inst. \\
\midrule
DeepSeek-V3.2 & EN & 0.864 & 0.850 & 0.857 & 5.5 \\
DeepSeek-V3.2 & ES & 0.906 & 0.728 & 0.808 & 4.7 \\
DeepSeek-V3.2 & PT & 0.824 & 0.407 & 0.544 & 2.8 \\
gemini-2.5-pro & EN & 0.911 & 0.485 & 0.633 & 3.0 \\
gemini-2.5-pro & ES & 0.915 & 0.458 & 0.611 & 2.9 \\
gemini-2.5-pro & PT & 0.938 & 0.522 & 0.670 & 3.2 \\
gpt-5.4-mini & EN & 0.931 & 0.498 & 0.649 & 3.1 \\
gpt-5.4-mini & ES & 0.927 & 0.488 & 0.640 & 3.1 \\
gpt-5.4-mini & PT & 0.947 & 0.667 & 0.782 & 4.2 \\
gpt-5.5 & EN & 0.941 & 0.510 & 0.661 & 3.2 \\
gpt-5.5 & ES & 0.922 & 0.547 & 0.686 & 3.4 \\
gpt-5.5 & PT & 0.937 & 0.503 & 0.655 & 3.1 \\
grok-4.3 & EN & 0.945 & 0.508 & 0.661 & 3.1 \\
grok-4.3 & ES & 0.941 & 0.477 & 0.633 & 3.0 \\
grok-4.3 & PT & 0.975 & 0.503 & 0.664 & 3.1 \\
\bottomrule
\end{tabular}
\caption{Second domain (weather, NOAA forecasts; complete record oracle). The effect replicates outside F1: the most precise model is not the most complete, so precision and $F_1$ disagree on the ranking. The effect is milder than in F1, as a weather record has fewer facts to omit.}
\label{tab:weather}
\end{table*}
}{}
\IfFileExists{models_table.tex}{\begin{table}[t]
\centering\small
\resizebox{\columnwidth}{!}{%
\begin{tabular}{lcccc}
\toprule
System (EN) & Prec. & Recall & F1 & Cl./inst. \\
\midrule
Qwen2.5-3B (zero-shot) & 0.825 & 0.666 & 0.737 & 7.0 \\
Qwen2.5-3B (fine-tuned) & 0.995 & 0.968 & 0.982 & 5.6 \\
\bottomrule
\end{tabular}}
\caption{Open small model (Qwen2.5-3B) zero-shot vs.\ LoRA fine-tuning on grounded explanations, held-out 2025 test sample, same precision+recall metric. Fine-tuning yields a model that is both \emph{accurate} and \emph{complete} (highest F1 in the study), reproducing the deterministic grounded templates -- a strength on this distribution and a template-mimicry caveat off it. No test leakage (\FOneFirstSeason{}--2024).}
\label{tab:models}
\end{table}
}{}
\IfFileExists{extractor_agreement.tex}{% Auto-generated by experiments/extractor_agreement.py
\begin{table}[t]
\centering
\small
\begin{tabular}{lcc}
\toprule
Model & Regex (EN) & LLM (EN) \\
\midrule
DeepSeek-V3.2 & 0.735 & 0.838 \\
gemini-2.5-pro & 0.925 & 0.901 \\
gpt-5.4-mini & 0.742 & 0.819 \\
gpt-5.5 & 0.753 & 0.843 \\
grok-4.3 & 0.749 & 0.855 \\
\bottomrule
\end{tabular}
\caption{English faithfulness under the model-free regex extractor vs.\ the LLM extractor (gpt-5.x). The two extractors agree on the system-level ranking (Spearman $0.8$; same top model) and correlate at the instance level (Pearson $0.4984$, $N=564$), showing the English results are not an artifact of the LLM extractor's family. The regex extractor's ES/PT patterns are deliberately light, so it serves as an English-first cross-check.}
\label{tab:extractor}
\end{table}
}{}

\section{Is It Just Prompting?}\label{sec:ablation}
A natural objection is that low coverage is an artifact of under-prompting. Our default
prompt is deliberately neutral (it asks the model to explain using only the data, with no
length instruction); we test the objection directly with an explicit \emph{cover-all}
prompt that asks the model to state every supportable fact (pit laps, compounds, stops,
the move and its outcome, time gained, positions). Table~\ref{tab:ablation} compares the
two. Asking for completeness \emph{does not} close the gap: mean recall does not rise --
it falls, $\AblMeanRecallA{}\!\to\!\AblMeanRecallB{}$, and only $\AblNUp{}$ of
$\AblNModels{}$ models improve at all. The extra verbosity does not add the facts that
mattered (and the added precision cost is the trade-off, made explicit). The low coverage
is therefore \emph{not} a prompting artifact, and a single-axis faithfulness score reports
none of this swing; precision and recall together do.

\section{Generalization: A Second Oracle}\label{sec:weather}
The abstention problem is a claim about reference-free precision metrics in general, not
about F1. We replicate it in a second domain with a complete oracle: public-domain NOAA
weather forecasts, where each record (temperature, wind, precipitation chance, sky) is the
enumerable fact set a forecast should cover. We generate forecasts grounded in $150$
records per language with the same frontier models and score them with the same
precision/recall machinery (Table~\ref{tab:weather}). The effect reappears: the most
precise model is again \emph{not} the most complete, so ranking by precision and by $F_1$
disagree. The effect is milder than in F1 -- a weather record has only a handful of facts,
so there is less to omit -- which is itself informative: the coverage penalty scales with
how much a faithful answer \emph{should} contain. That the precision/coverage gap appears
in two unrelated complete-oracle domains indicates it is a property of the precision-only
metric, not of any one dataset.

\section{Ethics and Licensing}\label{sec:ethics}
F1/FOM timing data carries usage restrictions; we release only code, derived
structured data, and annotations, not raw broadcast/telemetry feeds. The
benchmark concerns strategy explanation, not betting or outcome prediction.

\section{Limitations}
Our recall metric measures coverage of the facts our deterministic extractor derives;
these are high-precision but not exhaustive (event detection uses heuristics), so recall
is defined relative to this enumerable fact set rather than to every conceivable salient
detail -- the completeness we rely on is completeness \emph{of the derived oracle}. The
abstention finding is a property of reference-free precision metrics in general; we
demonstrate and replicate it in two unrelated complete-oracle domains (F1 and weather).
The claim extractor can miss or over-segment claims, so absolute scores should be read
alongside the controlled-perturbation validation. More fundamentally, the verifier only
checks the fact \emph{types} our schema models: ungrounded \emph{entity} or \emph{causal}
insertions outside the schema are never extracted, hence never penalised. For example,
asked to explain a two-lap defensive hold whose context names only the two drivers and a
\texttt{teammate\_protected} flag, one model named the protected teammate
(``Verstappen'') and credited it with a title-fight swing -- neither present in the given
context -- yet scored a perfect supported fraction. Faithfulness coverage is thus bounded
by the claim ontology; broadening it is future work. We mitigate the separate concern that
the gpt-5.x extractor shares a family with gpt-5.5 with a model-free regex extractor and a
cross-family LLM extractor that agree closely with it (Section~\ref{sec:exp}), but a
human extractor/judge remains future work. Batched LLM-extractor scoring is itself a
measurement risk: under API rate limiting, failed extractions can silently default to
zero, deflating coverage for later items in a batch and---in our own runs---momentarily
inverting a model ranking until we re-scored with retries and bounded concurrency. A low
coverage number can be a scoring artifact rather than a worse generation; coverage
evaluation must therefore guard its own measurement pipeline---our own thesis applied
reflexively. Fine-tuning supervision is silver
(deterministic faithful templates), which risks rewarding template mimicry; stronger
supervision and out-of-template evaluation are future work. The model comparison uses a
stratified test sample for cost, covers EN/ES/PT, and the held-out split is a single
season (2022 omitted: timing data unavailable). The two models served behind the hosted
AIServices endpoint had a fixed block of their English inputs ($\sim$one third) rejected
by the platform's default content filter -- the same instances for both models, so
input-triggered, not model behavior -- which we exclude from scoring and mark in
Table~\ref{tab:coverage}; the cleaner ES/PT cells, essentially unaffected, carry the same
conclusion. We also note that obtaining usable outputs from a reasoning model required
raising its output-token budget so that internal reasoning did not truncate the answer --
a reminder that the measurement pipeline, not only the model, must be audited.

Our headline model comparison focuses on the three core decision types (strategy,
undercut, overcut); we additionally evaluate the open-ended race-summary task
(Section~\ref{sec:exp}), while a full evaluation of the on-track defense type -- now
including the brief, overtake-ending holds the detector newly recovers -- is left to
future work. More broadly, requiring a \emph{complete} oracle limits applicability to
domains where the full fact set can be enumerated. A natural relaxation is a
\emph{retrieval-delta} metric: measuring the change in faithfulness when a model is given
grounding context versus generating from parametric knowledge alone. If the delta
correlates with oracle-measured recall (which our setting can validate), it would
generalize coverage-aware evaluation to open-domain settings without requiring
completeness---an avenue we leave to future work.

\bibliography{refs}

\begin{thebibliography}{21}
\providecommand{\natexlab}[1]{#1}

\bibitem[{Dhingra et~al.(2019)Dhingra, Faruqui, Parikh, Chang, Das, and
  Cohen}]{dhingra2019parent}
Bhuwan Dhingra, Manaal Faruqui, Ankur Parikh, Ming-Wei Chang, Dipanjan Das, and
  William Cohen. 2019.
\newblock Handling divergent reference texts when evaluating table-to-text
  generation.
\newblock In \emph{Proceedings of the 57th Annual Meeting of the Association
  for Computational Linguistics (ACL)}, pages 4884--4895.

\bibitem[{Es et~al.(2024)Es, James, Espinosa-Anke, and
  Schockaert}]{es2024ragas}
Shahul Es, Jithin James, Luis Espinosa-Anke, and Steven Schockaert. 2024.
\newblock {RAGAS}: Automated evaluation of retrieval augmented generation.
\newblock In \emph{Proceedings of the European Chapter of the Association for
  Computational Linguistics (EACL): System Demonstrations}.

\bibitem[{Fabbri et~al.(2022)Fabbri, Wu, Liu, and Xiong}]{fabbri2022qafacteval}
Alexander Fabbri, Chien-Sheng Wu, Wenhao Liu, and Caiming Xiong. 2022.
\newblock {QAFactEval}: Improved {QA}-based factual consistency evaluation for
  summarization.
\newblock In \emph{NAACL}.

\bibitem[{Gao et~al.(2023)Gao, Dai, Pasupat, Chen, Chaganty, Fan, Zhao, Lao,
  Lee, Juan, and Guu}]{gao2023rarr}
Luyu Gao, Zhuyun Dai, Panupong Pasupat, Anthony Chen, Arun~Tejasvi Chaganty,
  Yicheng Fan, Vincent Zhao, Ni~Lao, Hongrae Lee, Da-Cheng Juan, and Kelvin
  Guu. 2023.
\newblock {RARR}: Researching and revising what language models say, using
  language models.
\newblock In \emph{Proceedings of the 61st Annual Meeting of the Association
  for Computational Linguistics (ACL)}.

\bibitem[{Honovich et~al.(2022)Honovich, Aharoni, Herzig, Taitelbaum,
  Kukliansy, Cohen, Scialom, Szpektor, Hassidim, and Matias}]{honovich2022true}
Or~Honovich, Roee Aharoni, Jonathan Herzig, Hagai Taitelbaum, Doron Kukliansy,
  Vered Cohen, Thomas Scialom, Idan Szpektor, Avinatan Hassidim, and Yossi
  Matias. 2022.
\newblock {TRUE}: Re-evaluating factual consistency evaluation.
\newblock In \emph{Proceedings of the Conference of the North American Chapter
  of the Association for Computational Linguistics (NAACL)}.

\bibitem[{Ji et~al.(2023)Ji, Lee, Frieske, Yu, Su, Xu, Ishii, Bang, Madotto,
  and Fung}]{ji2023survey}
Ziwei Ji, Nayeon Lee, Rita Frieske, Tiezheng Yu, Dan Su, Yan Xu, Etsuko Ishii,
  Yejin Bang, Andrea Madotto, and Pascale Fung. 2023.
\newblock Survey of hallucination in natural language generation.
\newblock \emph{ACM Computing Surveys}.

\bibitem[{Kry{\'s}ci{\'n}ski et~al.(2020)Kry{\'s}ci{\'n}ski, McCann, Xiong, and
  Socher}]{kryscinski2020factcc}
Wojciech Kry{\'s}ci{\'n}ski, Bryan McCann, Caiming Xiong, and Richard Socher.
  2020.
\newblock Evaluating the factual consistency of abstractive text summarization.
\newblock In \emph{EMNLP}.

\bibitem[{Laban et~al.(2022)Laban, Schnabel, Bennett, and
  Hearst}]{laban2022summac}
Philippe Laban, Tobias Schnabel, Paul~N. Bennett, and Marti~A. Hearst. 2022.
\newblock {SummaC}: Re-visiting {NLI}-based models for inconsistency detection
  in summarization.
\newblock In \emph{TACL}.

\bibitem[{Liang et~al.(2009)Liang, Jordan, and Klein}]{liang2009weather}
Percy Liang, Michael~I. Jordan, and Dan Klein. 2009.
\newblock Learning semantic correspondences with less supervision.
\newblock In \emph{Proceedings of ACL-IJCNLP}, pages 91--99.

\bibitem[{Madaan et~al.(2023)Madaan, Tandon, Gupta
  et~al.}]{madaan2023selfrefine}
Aman Madaan, Niket Tandon, Prakhar Gupta, and 1 others. 2023.
\newblock Self-refine: Iterative refinement with self-feedback.
\newblock In \emph{NeurIPS}.

\bibitem[{Manakul et~al.(2023)Manakul, Liusie, and
  Gales}]{manakul2023selfcheckgpt}
Potsawee Manakul, Adian Liusie, and Mark~JF Gales. 2023.
\newblock {SelfCheckGPT}: Zero-resource black-box hallucination detection for
  generative large language models.
\newblock In \emph{Proceedings of the Conference on Empirical Methods in
  Natural Language Processing (EMNLP)}.

\bibitem[{Maynez et~al.(2020)Maynez, Narayan, Bohnet, and
  McDonald}]{maynez2020faithfulness}
Joshua Maynez, Shashi Narayan, Bernd Bohnet, and Ryan McDonald. 2020.
\newblock On faithfulness and factuality in abstractive summarization.
\newblock In \emph{Proceedings of the 58th Annual Meeting of the Association
  for Computational Linguistics (ACL)}, pages 1906--1919.

\bibitem[{Min et~al.(2023)Min, Krishna, Lyu, Lewis, Yih, Koh, Iyyer,
  Zettlemoyer, and Hajishirzi}]{min2023factscore}
Sewon Min, Kalpesh Krishna, Xinxi Lyu, Mike Lewis, Wen-tau Yih, Pang~Wei Koh,
  Mohit Iyyer, Luke Zettlemoyer, and Hannaneh Hajishirzi. 2023.
\newblock {FActScore}: Fine-grained atomic evaluation of factual precision in
  long form text generation.
\newblock In \emph{EMNLP}.

\bibitem[{Oehrly and contributors(2024)}]{fastf1}
Philipp Oehrly and contributors. 2024.
\newblock {FastF1}: Python package for accessing formula 1 timing and telemetry
  data.
\newblock \url{https://github.com/theOehrly/Fast-F1}.

\bibitem[{Parikh et~al.(2020)Parikh, Wang, Gehrmann, Faruqui, Dhingra, Yang,
  and Das}]{parikh2020totto}
Ankur Parikh, Xuezhi Wang, Sebastian Gehrmann, Manaal Faruqui, Bhuwan Dhingra,
  Diyi Yang, and Dipanjan Das. 2020.
\newblock {ToTTo}: A controlled table-to-text generation dataset.
\newblock \emph{EMNLP}.

\bibitem[{Santillana(2026)}]{vectrayxnano2026}
Juan~S. Santillana. 2026.
\newblock \href {https://arxiv.org/abs/2605.13989} {{VectraYX-Nano}: A
  42{M}-parameter {Spanish} cybersecurity language model with curriculum
  learning and native tool use}.
\newblock \emph{Preprint}, arXiv:2605.13989.

\bibitem[{Team(2025)}]{qwen2025}
Qwen Team. 2025.
\newblock Qwen2.5 technical report.
\newblock \emph{arXiv preprint}.

\bibitem[{Thomson and Reiter(2020)}]{thomson2020accuracy}
Craig Thomson and Ehud Reiter. 2020.
\newblock A gold standard methodology for evaluating accuracy in data-to-text
  systems.
\newblock In \emph{Proceedings of the 13th International Conference on Natural
  Language Generation (INLG)}, pages 158--168.

\bibitem[{Thomson et~al.(2020)Thomson, Reiter, and
  Sripada}]{thomson2020sportsett}
Craig Thomson, Ehud Reiter, and Somayajulu Sripada. 2020.
\newblock {SportSett}:{Basketball} -- a robust and maintainable data-set for
  natural language generation.
\newblock In \emph{Workshop on Intelligent Information Processing and Natural
  Language Generation}.

\bibitem[{Wei et~al.(2024)Wei, Yang, Song, Lu, Hou, Zhou, and
  Le}]{wei2024longfact}
Jerry Wei, Chengrun Yang, Xinying Song, Yifeng Lu, Le~Hou, Denny Zhou, and
  Quoc~V. Le. 2024.
\newblock Long-form factuality in large language models.
\newblock \emph{Advances in Neural Information Processing Systems (NeurIPS)}.

\bibitem[{Wiseman et~al.(2017)Wiseman, Shieber, and
  Rush}]{wiseman2017challenges}
Sam Wiseman, Stuart Shieber, and Alexander Rush. 2017.
\newblock Challenges in data-to-document generation.
\newblock In \emph{EMNLP}.

\end{thebibliography}

\end{document}